\documentclass[journal,twoside,web]{ieeecolor}
\usepackage{tmi}
\usepackage{cite}
\usepackage{amsmath,amssymb,amsfonts}
\usepackage{algorithmic}
\usepackage{graphicx}
\usepackage{textcomp}

\usepackage{url}
\usepackage{color}
\usepackage{subfigure}
\usepackage{graphicx}
\usepackage{multirow}
\usepackage{booktabs} 
\usepackage[marginal]{footmisc} 
\usepackage{adjustbox}
\usepackage{tipa} 
\usepackage{float} 
\usepackage{placeins}
\usepackage{bm} 
\usepackage{array} 
\usepackage{enumerate}
\usepackage{hyperref}
\usepackage{pifont} 
\usepackage{subfloat}
\usepackage{subfigure}
\usepackage{utfsym}
\usepackage{amsmath,bm}

\usepackage{diagbox}
\usepackage{booktabs}
\usepackage{amssymb}
\usepackage{balance} 
\usepackage{bbm} 

\usepackage{amsmath}

\def\BibTeX{{\rm B\kern-.05em{\sc i\kern-.025em b}\kern-.08em
    T\kern-.1667em\lower.7ex\hbox{E}\kern-.125emX}}
\begin{document}

\title{LapFM: A Laparoscopic Segmentation Foundation Model via Hierarchical Concept Evolving Pre-training}
\author{Qing Xu, Kun Yuan, Yuxiang Luo, Yuhao Zhai, Wenting Duan, Nassir Navab,
\IEEEmembership{Fellow, IEEE}, Zhen Chen
\thanks{\quad This work was supported in part by the Program of China Scholarship Council (202508330191). \textit{(Equal contribution: Q.
Xu, Y. Kun and Y. Luo, Corresponding author: Z. Chen)}} 
\thanks{\quad Q. Xu is with the School of Computer Science, University of Lincoln, UK, with University of Nottingham, UK, and with University of Nottingham Ningbo China, China (e-mail: qing.xu@nottingham.edu.cn).} 
\thanks{\quad K. Yuan is with University of Strasbourg, France, and with Technical University of Munich, Germany (e-mail: kun.yuan@ext.ihu-strasbourg.eu).} 
\thanks{\quad Y. Luo is with the Graduate School of Information, Production and Systems, Waseda University, Japan (e-mail: yuxiang.luo@ruri.waseda.jp).}
\thanks{\quad Y. Zhai is with Department of Gastrointestinal Surgery, The Second Qilu Hospital, Shandong University, China (e-mail: zhaiyuhao@email.sdu.edu.cn).}
\thanks{\quad W. Duan is with School of Engineering and Physical Science, University of Lincoln, Lincoln  LN6 7TS, UK (e-mail: wduan@lincoln.ac.uk).}
\thanks{\quad N. Navab is with Technical University of Munich, Germany (e-mail: nassir.navab@tum.de).}
\thanks{\quad Z. Chen is with Yale University, New Haven, CT 06510, USA (e-mail: zchen.francis@gmail.com).}
}

\maketitle
\begin{abstract}
Surgical segmentation is pivotal for scene understanding yet remains hindered by annotation scarcity and semantic inconsistency across diverse procedures. Existing approaches typically fine-tune natural foundation models (\textit{e.g.}, SAM) with limited supervision, functioning merely as domain adapters rather than surgical foundation models. Consequently, they struggle to generalize across the vast variability of surgical targets. To bridge this gap, we present LapFM, a foundation model designed to evolve robust segmentation capabilities from massive unlabeled surgical images. Distinct from medical foundation models relying on inefficient self-supervised proxy tasks, LapFM leverages a Hierarchical Concept Evolving Pre-training paradigm. First, we establish a Laparoscopic Concept Hierarchy (LCH) via a hierarchical mask decoder with parent-child query embeddings, unifying diverse entities (\textit{i.e.}, Anatomy, Tissue, and Instrument) into a scalable knowledge structure with cross-granularity semantic consistency. Second, we propose a Confidence-driven Evolving Labeling that iteratively generates and filters pseudo-labels based on hierarchical consistency, progressively incorporating reliable samples from unlabeled images into training. This process yields LapBench-114K, a large-scale benchmark comprising 114K image-mask pairs. Extensive experiments demonstrate that LapFM significantly outperforms state-of-the-art methods, establishing new standards for granularity-adaptive generalization in universal laparoscopic segmentation. The source code is available at \url{https://github.com/xq141839/LapFM}.
\end{abstract}


\section{Introduction}
\label{sec:introduction}
\IEEEPARstart{S}{urgical} segmentation is fundamental to advancing computer-assisted intervention systems, providing critical visual understanding for intraoperative navigation, risk assessment, and robotic automation \cite{twinanda2016endonet,kolbinger2023anatomy}. While deep learning has achieved remarkable success in specific tasks \cite{ronneberger2015u,chen2024transunet,maack2024efficient, tomar2025effective,chen2023surgnet,yang2023msde,zhao2025rethinking}, the fragmented nature of surgical data, varying widely across procedures, instruments, and patient anatomies, calls for foundation models capable of universal scene understanding. Such foundation models should robustly identify diverse entities, from anatomical structures to surgical instruments, without requiring retraining for every new scenario \cite{carstens2023dresden,hong2020cholecseg8k,wang2022autolaparo}.


The recent advance of Segment Anything Model (SAM) \cite{kirillov2023segment, ravisam} has inspired a wave of surgical SAM variants \cite{yue2024surgicalsam, liusurgical, kamtam2025surgisam2, liu2025resurgsam2, sheng2024surgical, huang2024segment,wu2025medical,yan2025samed, xu2025lightsam}. However, most existing works are limited by a \textit{fixed-granularity fine-tuning} paradigm. While they often utilize parameter-efficient techniques (\textit{e.g.}, adapters) to transfer natural image knowledge, their fundamental limitation lies in the learning objective \cite{yue2024surgicalsam,liusurgical, Shaharbany_2023_BMVC, cheng2024unleashing, ward2025autoadaptive}. These methods rely on the fixed, pre-defined categories of small-scale supervised datasets and lack a unified semantic framework to comprehend the hierarchical relationships in complex surgical scenes. Consequently, even with adapters, they fail to generalize when annotation granularities differ (\textit{e.g.}, \textit{tool} vs. \textit{grasper}) or when facing unseen surgical categories \cite{chen2023surgnet, yang2023msde, zhao2025rethinking}.

\begin{figure*}[!t]
  \centering
  \includegraphics[width=0.95\linewidth]{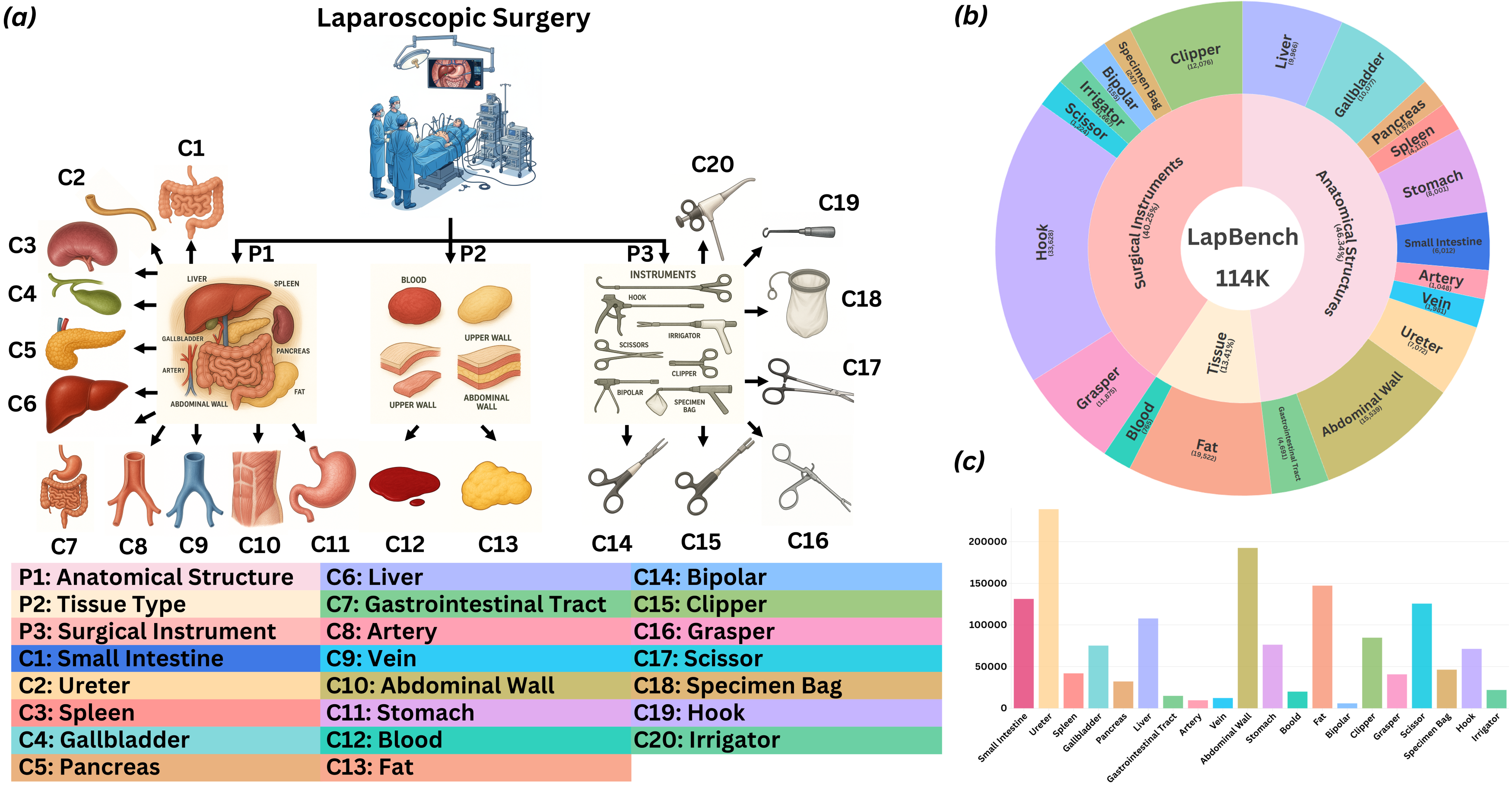}
\caption{(a) The LCH that unifies diverse surgical entities into a scalable taxonomy with three fundamental branches: \textit{Anatomy}, \textit{Tissue}, and \textit{Instrument}. Parent (P) nodes guide Child (C) nodes for granularity-adaptive segmentation from coarse to fine levels. 
(b) Distribution of segmentation labels per image in our LapBench-114K dataset. 
(c) Histogram of pixel-level annotations across surgical categories.}
  \label{fig:intro}
\end{figure*}

Constructing a surgical foundation model faces a unique challenge regarding data utilization. Despite the abundance of raw surgical images \cite{twinanda2016endonet,nwoye2022rendezvous}, obtaining pixel-level semantic annotations is exceptionally resource-intensive. While traditional medical foundation models attempt to bypass this by adopting Self-Supervised Learning (SSL) techniques like masked autoencoders or contrastive learning to leverage unlabeled data \cite{zhou2021models,ye2023uniseg}, these generic proxy tasks are ill-suited for surgical segmentation tasks. They tend to learn global representations, rather than the fine-grained, pixel-level features necessary to parse complex semantic structures like instrument-tissue interactions inherent in surgical scenes.


To address these bottlenecks, we propose LapFM, a Laparoscopic Foundation Model built via a Hierarchical Concept Evolving Pre-training paradigm. Our core insight is that a foundation model should not just learn from static data but \textit{evolve} by actively exploring the unlabeled world. First, to resolve the semantic inconsistency across datasets \cite{carstens2023dresden,hong2020cholecseg8k,murali2023endoscapes}, we construct a Laparoscopic Concept Hierarchy (LCH), as illustrated in Fig. \ref{fig:intro}(a). This structure maps fragmented surgical categories into a unified taxonomy rooted in three fundamental branches: \textit{Anatomy}, \textit{Tissue}, and \textit{Instrument}. To establish this hierarchy, LapFM utilizes a hierarchical mask decoder with specialized parent and child query embeddings, where parent-specific features guide child-level segmentation, explicitly enforcing semantic consistency across granularity levels. This allows LapFM to bridge coarse-grained and fine-grained annotations within a hierarchical framework. Moreover, distinct from inefficient SSL proxy tasks, we propose a Confidence-driven Evolving Labeling. We formulate the segmentation task as the learning objective for unlabeled data, iteratively generating pseudo-labels and filtering them based on confidence checks. This process effectively transforms massive unlabeled data into high-quality supervision knowledge, enabling the LapFM to learn robust representations from diverse surgical scenes. We validate LapBench-114K through random sampling by a certified surgeon, confirming the reliability of the generated pseudo-labels for LapFM learning.


The contributions of this work are summarized as follows:
\begin{itemize}

\item We propose LapFM, a laparoscopic foundation model built via Hierarchical Concept Evolving Pre-training, designed to evolve robust segmentation capabilities from massive unlabeled surgical images.

\item We devise a LCH that unifies diverse surgical entities into a scalable knowledge structure with cross-granularity semantic consistency, and propose a Confidence-driven Evolving Labeling that iteratively generates reliable pseudo-labels from unlabeled data.

\item We present LapBench-114K, a large-scale laparoscopic surgical segmentation benchmark for comprehensive evaluation, which is generated progressively with high-quality masks in varying surgical categories and granularities.

\item Extensive experiments prove that our LapFM surpasses state-of-the-art segmentation methods, with remarkable generalization capabilities.
\end{itemize}

\begin{figure*}[!t]
  \centering
  \includegraphics[width=0.95\linewidth]{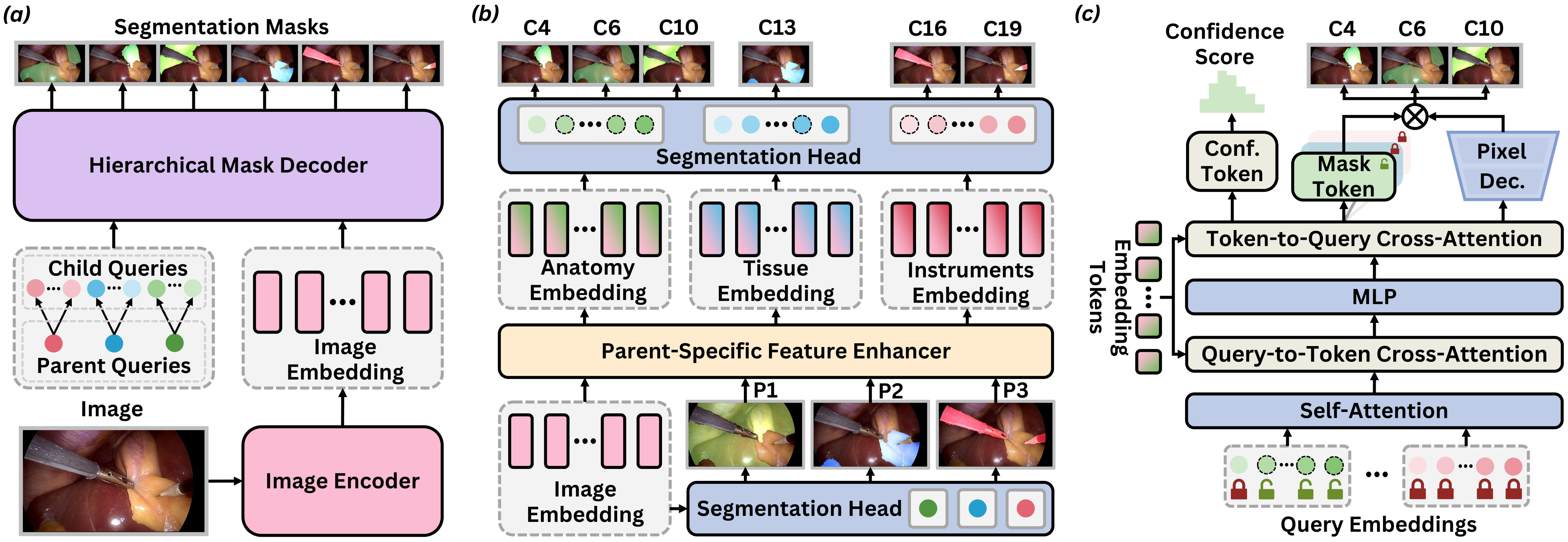}
\caption{The overview of the proposed LapFM framework, consisting of (a) a transformer-based image encoder and a hierarchical mask decoder for multi-granularity segmentation. 
(b) The hierarchical mask decoder leverages parent and child query embeddings with explicit parent-child dependencies, where parent-specific features guide child-level concept segmentation. 
(c) Detailed architecture of the segmentation head. We illustrate anatomy segmentation as an example to demonstrate how LapFM achieves adaptive hierarchy traversal across granularity levels.}
  \label{fig:method}
\end{figure*}

\section{Related Work}
\subsection{Surgical Segmentation}

Surgical procedures have evolved significantly from traditional open surgery to advanced minimally invasive techniques, encompassing laparoscopic surgery \cite{wang2023sam, yue2024surgicalsam, nasirihaghighi2025gynsurg}. Among these, laparoscopic surgery has become the gold standard for numerous abdominal procedures due to its reduced trauma, faster recovery, and improved patient outcomes. However, the constrained field of view, limited depth perception, and complex tissue-instrument interactions in laparoscopic environments pose significant challenges for surgeons. In this context, surgical segmentation emerges as a fundamental technology for computer-assisted intervention systems, enabling precise delineation of surgical instruments, anatomical structures, and tissue types to facilitate surgical navigation, risk assessment, and automated decision support \cite{kolbinger2023anatomy}. Classical studies \cite{ chen2024transunet, liu2024swin, chen2025zig} utilized the U-Net \cite{ronneberger2015u} architecture to achieve automatic surgical segmentation.

In the realm of surgical instrument segmentation that aims to identify individual tools and their functional parts \cite{wang2022autolaparo}, existing studies \cite{ni2022surginet, yang2022tmf, yang2023msde} have made notable strides through tailored network architectures. SurgCSS \cite{zhao2025rethinking} leveraged class-aware blending and edge bias correction to address imbalanced surgical instrument segmentation. For anatomical structure segmentation, MT-KD \cite{maack2024efficient} adopted multi-teacher knowledge distillation to achieve real-time segmentation while maintaining high accuracy. Moreover, EDRL \cite{tomar2025effective} alleviated representation bias in surgical multi-organ segmentation through pre-training and contextual feature sharing between decoders. Despite the progress, these methods focused on specific surgical segmentation tasks, struggling to handle the inherent variability of real surgical scenarios, including tissue occlusion and image degradation. In contrast, our LapFM unifies diverse surgical entities into a scalable knowledge structure, enabling granularity-adaptive segmentation and robust performance across diverse laparoscopic surgical scenarios.

\subsection{Foundation Models for Medical Segmentation}
The classical medical foundation segmentation frameworks have emerged to address the challenge of generalizing across diverse medical imaging modalities and anatomical structures. Models Genesis \cite{zhou2021models} pioneered self-supervised pre-training on large-scale unlabeled medical images, establishing transferable representations for downstream segmentation tasks across different modalities. Building upon this basis, UniSeg \cite{ye2023uniseg} utilized shared feature representations and category-specific decoders to segment multiple anatomical structures.

Recent advances in prompt-driven segmentation have introduced more flexible paradigms for medical image analysis. The SAM series \cite{kirillov2023segment, ravisam} revolutionized segmentation through interactive prompting mechanisms, enabling generalization across medical domains by subsequent adaptations \cite{wang2023sam, sheng2024surgical, ma2024segment, huang2024segment,wu2025medical,yan2025samed,kamtam2025surgisam2, xu2025lightsam}. In particular, SurgicalSAM \cite{yue2024surgicalsam} addressed the domain gap in surgical instruments through a prototype-based class prompt encoder with contrastive learning. SurgSAM-2 \cite{liusurgical} developed a selective memory management for faster surgical video inference. ASI-Seg \cite{chen2024asi} introduced an audio-driven segmentation framework that interprets surgeons' verbal commands through multimodal fusion, enabling intention-oriented instrument segmentation. In addition, CLIP-driven approaches \cite{luddecke2022image, liu2023clip} further extended this paradigm by incorporating text-based semantic guidance, enabling category-agnostic segmentation through natural language descriptions. For example, ReSurgSAM2 \cite{liu2025resurgsam2} combined cross-modal spatial-temporal Mamba with diversity-driven memory for text-referred surgical segmentation. Despite the progress, these methods rely on predefined categorical spaces established during training, limiting their adaptability to varying annotation granularities and emerging surgical categories in clinical practice. Different from these approaches, our LapFM unifies diverse surgical entities into a scalable knowledge structure, achieving fully granularity-adaptive segmentation without requiring manual prompts.

\section{Methodology}

We present the LapFM framework based on Hierarchical Concept Evolving Pre-training in Fig. \ref{fig:method}. The core of our approach consists of two synergistic components: (1) a LCH that unifies diverse surgical entities into a scalable knowledge structure, and (2) a Confidence-driven Evolving Labeling that iteratively expands the knowledge of LapFM from high-confidence seeds to complex long-tail scenarios, ultimately constructing the large-scale LapBench-114K benchmark for a comprehensive understanding of laparoscopic surgery.

\subsection{Laparoscopic Concept Hierarchy}
Existing surgical segmentation methods \cite{yue2024surgicalsam, liusurgical, liu2025resurgsam2} rely on predefined category spaces specific to individual datasets,  failing to generalize when confronted with different categories or annotation granularities across diverse surgical scenarios. To resolve this semantic inconsistency, we construct a LCH that maps fragmented surgical categories into a unified taxonomy rooted in three fundamental branches: \textit{Anatomy}, \textit{Tissue}, and \textit{Instrument}. This structure establishes semantic relationships across granularity levels and enables dynamic category matching for seamless adaptation to emerging laparoscopic targets across diverse surgical scenarios.

\begin{figure}[!t]
  \centering
  \includegraphics[width=1\linewidth]{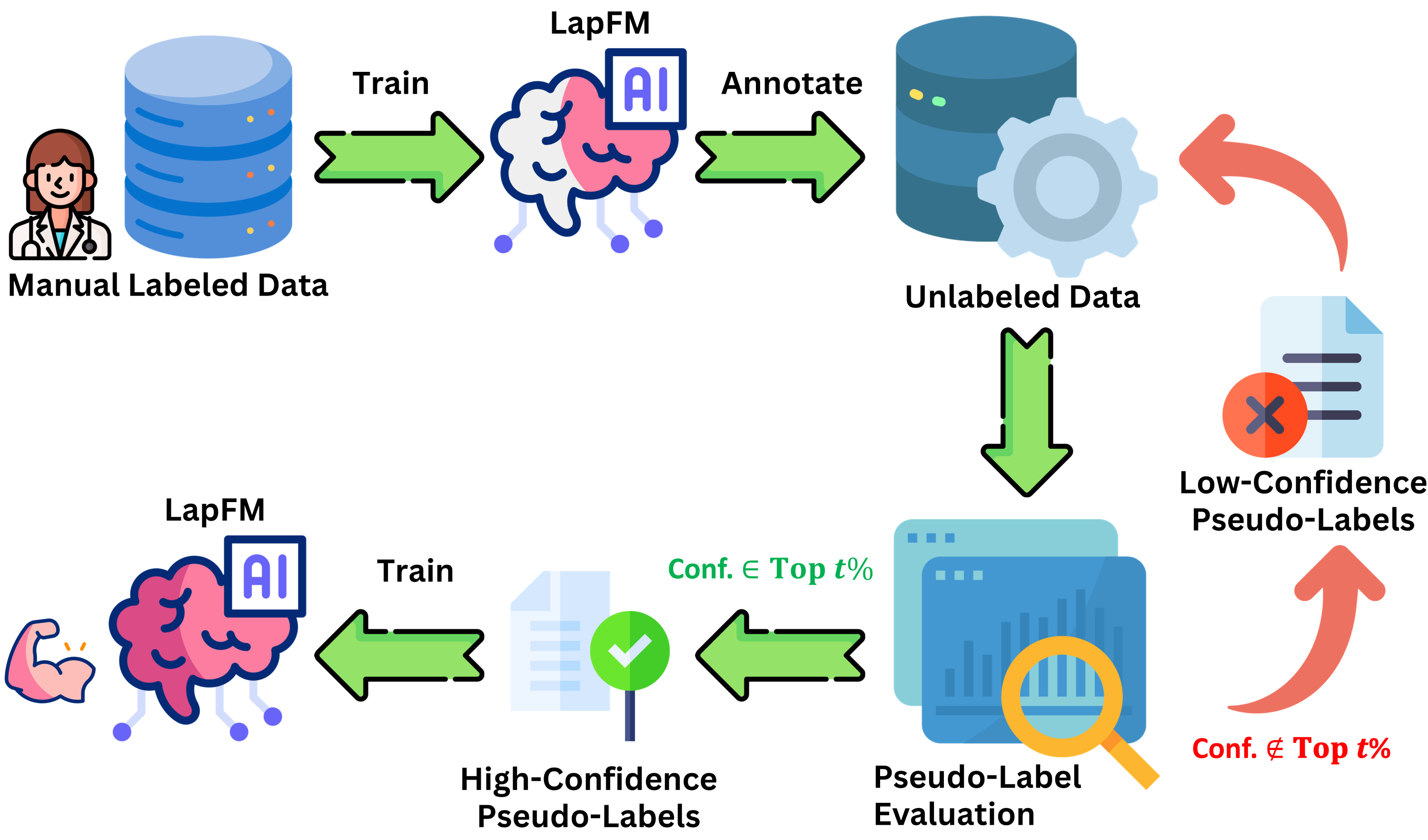}
  \caption{The overview of our Confidence-driven Evolving Labeling. This process exploits model-assisted pseudo-labeling and confidence filtering to maximize data utilization while ensuring high-quality annotations.}
  \label{fig:loop}
\end{figure} 

Specifically, to comprehend the complex hierarchical relationships in surgical scenes and enable granularity-adaptive segmentation, our LCH explicitly models parent-child dependencies to facilitate knowledge transfer across granularity levels. To this end, we define the hierarchy structure $T = (N, E)$, where $N$ represents the set of concept nodes and the edge set $E$ identifies parent-child relationships between nodes:
\begin{equation} \label{eq:tree}
N=\bigcup_{l=0}^{L} V_l, \quad E = \{(v_i, v_j) | v_i \in V_{l}, v_j \in V_{l+1}\},
\end{equation}
where $L$ is the LCH depth, and $V_l$ represents the set of nodes at level $l$. The root level ($l=0$) contains three fundamental branches (\textit{Anatomy}, \textit{Tissue}, \textit{Instrument}), while deeper levels progressively refine these concepts into fine-grained surgical entities. The hierarchical prediction probability for the segmentation mask $y$ is formulated as:
\begin{equation} \label{eq:hier_prob}
p_\theta(y| x) = \prod_{l=1}^{L-1} p_\theta(V_{l+1} | V_l, x),
\end{equation}
where $p_\theta(V_{l+1} | V_l, x)$ implies that knowing the parent concept segmentation $V_l$ affects the outcome of the child concept segmentation $V_{l+1}$. Therefore, the Eq. \eqref{eq:hier_prob} formulates dependencies between hierarchical nodes through conditional probabilities, ensuring that:
\begin{itemize}
  \item Fine-grained child concepts inherit semantic information from their coarse-grained parents, maintaining consistency across granularity levels.
  \item The prediction path from parent to child nodes enables adaptive granularity selection based on domain-specific requirements.
\end{itemize}
The gradient calculation for hierarchical concept learning is defined as:
\begin{equation} \label{eq:gradient}
\nabla_\theta \ell = - \mathbb{E} \left[ \nabla_\theta \log p_\theta(y | x) \right] -  \mathbb{E} \left[ \nabla_\theta \log p_\theta(V_{l+1} | V_l) \right],
\end{equation}
where the first term represents the main segmentation loss across target concepts, and the second term enforces prediction consistency between parent-child pairs. This hierarchical constraint ensures that semantically related concepts share similar feature representations, facilitating knowledge transfer across granularity levels. In this way, the proposed LCH bridges coarse-grained and fine-grained annotations within a single searchable framework, enabling LapFM to achieve diverse laparoscopic segmentation.

\begin{figure*}[!t]
  \centering
  \includegraphics[width=1\linewidth]{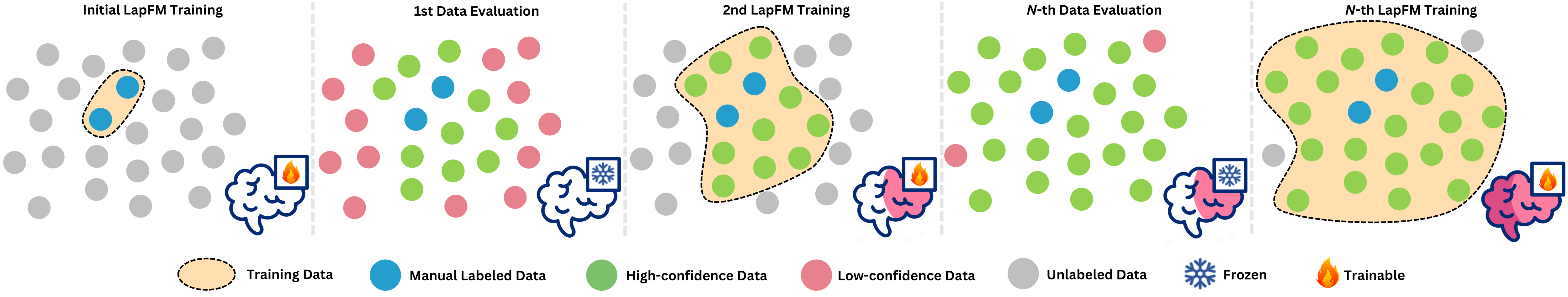}
  \caption{The progression of data expansion via the Confidence-driven Evolving Labeling. This process iteratively corrects low-confidence mask samples from reliable high-quality annotations and progressively integrates datasets with varying categories and granularities, exploiting cross-dataset surgical knowledge to construct LapBench-114K.}
  \label{fig:loop2}
\end{figure*} 

\subsection{The LapFM Architecture Designs}
Following the LCH, we design the LapFM architecture to enforce adaptive hierarchy traversal across different granularity levels, ensuring flexible concept retrieval rather than treating surgical categories as isolated scenes. LapFM adopts a transformer-based image encoder to process the input image $x$ for capturing long-range dependencies, generating image embeddings as $h=\varepsilon(x)$. To achieve universal segmentation decoding, we devise a hierarchical mask decoder for multi-granularity decoding. 

Specifically, for parent-level decoding, we set a group of parent query embeddings $q^l$. We first perform self-attention on parent queries, as $q^l \gets {\rm SA}(q^l)$ followed by bidirectional cross-attention with image embeddings. These interactions comprise a segmentation head that enables the insertion of decoding information into the image embedding for updating feature representations. The enhanced embeddings $h$ and $q^l$ are then respectively processed through a pixel decoder consisting of two $2 \times 2$ upsampling convolution layers, and parent MLPs as mask tokens (MT) to generate parent segmentation results:
\begin{equation}
y^l = {\rm MLP}(q^l) \cdot {\rm Upsample}(h).
\end{equation}
On this basis, we further set a group of child query embeddings $q^{l+1}$ for child-level segmentation decoding. To leverage the hierarchical relationships across different granularities, we design a parent-specific feature enhancer that performs self-attention on each encoded parent mask, as $y^l \gets {\rm Conv}(y^l)$, followed by cross-attention with the image embedding tokens:
\begin{equation}
    h^l = \mathrm{softmax}(\frac{(h+ \psi) \cdot (y^l)^{\top}}{\sqrt{d}})\cdot y^l+h,
\end{equation}
where $d$ is the channel dimension for the input embeddings, and $\psi$ represents the positional encoding. After that, we implement an additional segmentation head, as follows: 
\begin{equation}
\begin{split}
    h^l &\gets \mathrm{softmax}(\frac{(h^l+ \psi) \cdot (q^{l+1})^{T}}{\sqrt{d}})\cdot q^{l+1}+h, \\
    q^{l+1} &\gets \mathrm{softmax}(\frac{(q^{l+1}+ \psi) \cdot (h^l)^{T}}{\sqrt{d}})\cdot h^l+q^{l+1}.
\end{split}
\end{equation}
When parent-specific features act as keys and values in the attention mechanism, while child-level features serve as queries, this design allows the cross-attention operation to integrate coarse-grained contextual information into fine-grained features, enabling hierarchical constraints to guide concept-specific segmentation. For example, instrument sub-concepts (\textit{e.g.}, \textit{grasper} and \textit{hook}) benefit from the broader spatial context provided by their parent concept, \textit{i.e.}, surgical instruments. Similarly, we leverage the pixel decoder and corresponding child MLPs to produce child masks:
\begin{equation}
y^{l+1} = {\rm MLP}(q^{l+1}) \cdot {\rm Upsample}(h^l).
\end{equation}
To enforce consistency across different granularity levels of LCH, we apply Kullback-Leibler (KL) divergence to compute hierarchical consistency loss $\mathcal{L}_{\rm HC}$:
\begin{equation}
\mathcal{L}_{\rm HC} = \sum_{l=1}^{L} \mathbb{E}_{x \sim y^l_i(x)} \left[ \log y^l_i(x) -\log \max_{j \in C(i)} y_j^{l+1}(x) \right],
\end{equation}
where $C(i)$ denotes the set of child concepts belonging to parent $i$, and $\max_{j \in C(i)} y_j^{l+1}(x)$ represents the aggregated child probability. This constraint ensures that parent predictions align with the aggregation of their child predictions. For individual mask supervision, we utilize the combination of cross-entropy loss $\mathcal{L}_{\rm BCE}$ and Dice loss $\mathcal{L}_{\rm Dice}$ for each concept at both levels. Moreover, we create a set of small confidence tokens to predict the Dice score for each generated segmentation mask. The Dice prediction head is trained with mean-square-error loss $\mathcal{L}_{\rm MSE}$ between the Dice prediction and the predicted mask's Dice with the ground truth mask:
\begin{equation}
\mathcal{L}_{\rm MSE} = \sum_{l=1}^{L} \big|\big| t^l - \mathrm{Dice}(y^l, \hat{y}^l) \big|\big|^2.
\end{equation}
The overall loss of optimizing LapFM during the training stage is formulated as:
\begin{equation}
    \mathcal{L}_{\rm LapFM} = \mathcal{L}_{\rm Dice}+ \mathcal{L}_{\rm BCE} + \mathcal{L}_{\rm MSE} + \lambda_1\mathcal{L}_{\rm HC},
\end{equation}
where $\lambda_1$ is the coefficient to balance the hierarchical consistency constraint. In this way, LapFM leverages hierarchical mask decoding with explicit parent-child dependencies to achieve universal segmentation across diverse laparoscopic surgical domains with varying granularities.

\subsection{Confidence-driven Evolving Labeling}
While surgical images are abundant, pixel-level annotations remain scarce and expensive, creating a fundamental Data Paradox. Traditional foundation models \cite{zhou2021models,ye2023uniseg} attempt to bypass this through SSL techniques like masked autoencoders, which often learn global representations suboptimal for precision-demanding surgical segmentation. To address this challenge, we propose a Confidence-driven Evolving Labeling that transforms LapFM into a self-improving data engine, treating the segmentation task itself as the learning objective for unlabeled data, as illustrated in Fig. \ref{fig:loop}. The progression comprises two phases: initial training and iterative self-evolution. Specifically, we first collect existing publicly labeled laparoscopic surgical datasets with varying categories and annotation granularities, organizing them into our LCH spanning \textit{Anatomy}, \textit{Tissue}, and \textit{Instrument} branches. We train LapFM on this unified dataset, leveraging a pre-trained SAM2 \cite{ravisam} backbone as the image encoder for its strong visual representation capabilities. To reduce computation cost, we insert adapters into the FFN layers for parameter-efficient fine-tuning. The hierarchical mask decoder learns to handle surgical concepts with different granularities through the LCH structure.

After training the initial LapFM, we further collect additional unlabeled laparoscopic surgical images and employ LapFM to automatically generate pseudo-labels across all hierarchical levels. Distinct from inefficient SSL proxy tasks, this Confidence-Driven Evolving approach uses the segmentation task itself as the learning objective for unlabeled data. To ensure annotation quality, we adopt a confidence-guided iterative strategy that automatically identifies samples with high-quality masks. Specifically, for each iteration, we rank all predicted masks by their Dice confidence scores and select the top $t\%$ as high-confidence pseudo-labels, while the remaining samples are classified as low-confidence seeds. The high-confidence pseudo-labels are progressively integrated with the original labeled dataset for the next training iteration. For overlapping regions, we select pixels with maximum confidence scores. Through this confidence-driven mechanism, LapFM progressively expands from high-confidence seeds to complex long-tail scenarios, as visualized in Fig. \ref{fig:loop2}, effectively transforming massive unlabeled data into high-quality supervision.

\begin{table*}[!t]
\centering
\setlength\tabcolsep{8pt}
\caption{Comparison with state-of-the-arts on Anatomical Structure Segmentation.}
\adjustbox{width=1\textwidth,center}{\begin{tabular}{l|c|c|ccccccccccc}
\hline
Methods & MP & P1 & C1 & C2 & C3 & C4 & C5 & C6 & C7 & C8 & C9 & C10 & C11 \\
\hline
nnUnet \cite{isensee2021nnu} & \multirow{6}{*}{\usym{2715}} & 71.85 & 32.65 & 33.73 & 25.93 & 64.43 & 12.95 & 74.16 & 46.42 & 3.15 & 7.72 & 65.88 & 78.36 \\
Surginet \cite{ni2022surginet} & & 67.73 & 41.83 & 24.87 & 31.93 & 47.26 & 10.12 & 61.61 & 27.66 & 10.20 & 7.92 & 59.98 & 82.37 \\
MSDE-Net \cite{yang2023msde} &  & 65.72 & 18.89 & 55.14 & 12.38 & 50.53 & 1.43 & 57.72 & 38.62 & 1.17 & 3.54 & 61.93 & 79.14 \\
UniverSeg \cite{butoi2023universeg} & & 66.18 & 13.68 & 22.79 & 23.01 & 50.70 & 2.05 & 60.53 & 13.03 & 5.14 & 3.15 & 68.25 & 72.60 \\
SurgCSS \cite{zhao2025rethinking} & & 71.38 & 28.19 & 45.54 & 20.85 & 66.85 & 12.29 & 68.74 & 38.26 & 21.21 & 2.14 & 71.62 & 85.59 \\
Zig-RiR \cite{chen2025zig} & & 73.60 & 44.97 & 29.42 & 53.87 & 72.82 & 12.80 & 66.49 & 37.17 & 4.74 & 3.51 & 62.60 & 84.92 \\
\hline
SAM \cite{kirillov2023segment} & \multirow{6}{*}{\usym{1F5F8}} & 78.69 & 69.97 & 67.06 & 70.11 & 69.23 & 10.22 & 67.67 & 74.21 & 14.36 & 30.29 & 82.93 & 67.40 \\
SAM2 \cite{ravisam} &  & 80.45 & 72.38 & 69.52 & 72.84 & 71.67 & 11.54 & 70.23 & 76.89 & 17.62 & 34.71 & 84.36 & 70.18 \\
MedSAM \cite{ma2024segment} & & 81.23 & 73.15 & 70.84 & 74.92 & 73.48 & 12.37 & 71.68 & 79.64 & 21.47 & 38.26 & 83.57 & 72.93 \\
SurgicalSAM \cite{yue2024surgicalsam} & & 82.17 & 74.68 & 71.93 & 76.45 & 75.29 & 12.89 & 72.84 & 81.52 & 25.83 & 42.14 & 84.19 & 74.68 \\
SurgSAM-2 \cite{liusurgical} & & 83.28 & 75.84 & 73.15 & 78.93 & 77.16 & 13.28 & 73.76 & 85.47 & 30.52 & 46.35 & 84.58 & 76.54 \\
ReSurgSAM2 \cite{liu2025resurgsam2} & & 84.13 & 76.62 & \textbf{74.28} & 81.33 & 78.73 & 13.65 & 74.35 & \textbf{88.36} & 35.77 & 50.78 & 84.71 & 78.82 \\
\hline
AutoSAM \cite{Shaharbany_2023_BMVC} & \multirow{3}{*}{\usym{2715}} & 76.54 & 67.83 & 65.42 & 68.75 & 67.91 & 9.87 & 66.28 & 72.39 & 12.68 & 28.54 & 81.46 & 65.73 \\
H-SAM \cite{cheng2024unleashing} & & 79.62 & 71.24 & 68.37 & 72.59 & 71.85 & 11.73 & 69.54 & 76.28 & 19.43 & 36.82 & 83.28 & 69.47 \\
ADA-SAM \cite{ward2025autoadaptive} & & 82.37 & 74.95 & 71.68 & 76.82 & 75.64 & 12.94 & 72.93 & 82.76 & 27.35 & 44.68 & 84.82 & 75.29 \\
\hline
LapFM & \usym{2715} & \textbf{86.14} & \textbf{89.72} & 73.98 & \textbf{81.34} & \textbf{86.07} & \textbf{46.73} & \textbf{91.09} & 74.60 & \textbf{57.47} & \textbf{58.68} & \textbf{92.93} & \textbf{98.56} \\
\hline
\end{tabular}}
\label{tab:anatomy_dice}
\end{table*}

\begin{table}[!t]
\centering
\small
\setlength\tabcolsep{4pt}
\caption{Comparison with state-of-the-arts on Tissue Segmentation.}
\adjustbox{width=0.485\textwidth,center}{\begin{tabular}{l|c|c|cc|c|cc}
\hline
\multirow{2}{*}{Methods} & \multirow{2}{*}{MP} & \multicolumn{3}{c|}{Dice (\%) $\uparrow$} & \multicolumn{3}{c}{HD (mm) $\downarrow$} \\
\cline{3-8}
& & P2 & C12 & C13 & P2 & C12 & C13  \\
\hline
nnUnet \cite{isensee2021nnu} & \multirow{6}{*}{\usym{2715}} & 79.72 & 17.96 & 79.54 & 287.35 & 512.68 & 294.73 \\
Surginet \cite{ni2022surginet} &  & 74.95 & 12.83 & 75.24 & 324.87 & 578.92 & 331.56 \\
MSDE-Net \cite{yang2023msde} & & 76.66 & 13.45 & 76.92 & 308.54 & 543.37 & 315.68 \\
UniverSeg \cite{butoi2023universeg} & & 75.58 & 15.76 & 75.77 & 346.92 & 524.85 & 353.29 \\
SurgCSS \cite{zhao2025rethinking} & & 84.02 & 19.73 & 84.31 & 335.46 & 487.93 & 341.85 \\
Zig-RiR \cite{chen2025zig} & & 86.65 & 21.58 & 86.95 & 318.73 & 468.24 & 325.39 \\
\hline
SAM \cite{kirillov2023segment} & \multirow{6}{*}{\usym{1F5F8}}  & 75.33 & 7.82 & 75.47 & 342.77 & 122.10 & 465.67 \\
SAM2 \cite{ravisam} &  & 79.73 & 33.47 & 79.76 & 310.22 & 93.53 & 473.07 \\
MedSAM \cite{ma2024segment} & & 38.92 & 68.79 & 38.85 & 629.65 & 194.98 & 300.88 \\
SurgicalSAM \cite{yue2024surgicalsam} & & 81.47 & 38.65 & 81.53 & 295.84 & 87.26 & 448.35 \\
SurgSAM-2 \cite{liusurgical} & & 82.89 & 26.15 & 83.04 & 299.44 & 197.42 & 373.64 \\
ReSurgSAM2 \cite{liu2025resurgsam2} & & 85.21 & 48.75 & 85.22 & 283.88 & 80.35 & 284.40 \\
\hline
AutoSAM \cite{Shaharbany_2023_BMVC} & \multirow{3}{*}{\usym{2715}} & 73.28 & 6.54 & 73.39 & 367.59 & 645.73 & 489.24 \\
H-SAM \cite{cheng2024unleashing} & & 77.86 & 29.47 & 77.92 & 328.45 & 118.67 & 456.83 \\
ADA-SAM \cite{ward2025autoadaptive} & & 80.65 & 35.82 & 80.71 & 307.28 & 95.48 & 442.16 \\
\hline
LapFM & \usym{2715} & \textbf{92.96} & \textbf{98.41} & \textbf{92.95} & \textbf{279.45} & \textbf{9.72} & \textbf{280.02} \\
\hline
\end{tabular}}
\label{tab:tissue_comp}
\end{table}

\section{Experiments}

\subsection{Datasets and Implementations}

\subsubsection{Datasets}
To validate the effectiveness of the proposed LapFM, we conduct comprehensive evaluations on our constructed large-scale LapBench-114K dataset comprising 114K image-mask pairs across 20 surgical categories, including 11 anatomical structures, 2 tissue types, and 7 surgical instruments. Specifically, LapBench-114K integrates two types of data sources: (1) five fully-annotated datasets including Cholecseg8k \cite{hong2020cholecseg8k}, Dresden \cite{carstens2023dresden}, EndoScapes \cite{mascagni2025endoscapes}, M2caiSeg \cite{maqbool2020m2caiseg}, and AutoLaparoT3 \cite{wang2022autolaparo}, and (2) three unlabeled datasets including CholecT50 \cite{nwoye2022rendezvous}, Cholec80 \cite{twinanda2016endonet}, and Endoscapes-CVS201 \cite{murali2023endoscapes}, which are automatically annotated with high-quality pseudo-labels generated by our Confidence-driven Evolving Labeling. Moreover, we evaluate the generalization capability of LapFM on the unseen Gynsurg \cite{nasirihaghighi2025gynsurg} dataset.
 The details are as follows:

\noindent \textbf{CholecSeg8k} \cite{hong2020cholecseg8k} is a surgical segmentation dataset containing 8,080 laparoscopic images extracted from 17 cholecystectomy video clips. The dataset provides pixel-level annotations for 12 surgical categories with the resolution of $854 \times 480$.

\noindent \textbf{Dresden} \cite{carstens2023dresden} comprises 13,195 laparoscopic frames from 20 different surgeries across 11 anatomical structures, \textit{e.g.}, abdominal wall, liver, spleen, vein, ureter, and stomach. All images are stored at a resolution of $1,920 \times 1,080$.

\noindent \textbf{EndoScapes} \cite{mascagni2025endoscapes} contains 493 annotated frames from 50 laparoscopic cholecystectomy videos comprising 6 surgical categories. Each mask is annotated at regular intervals: 1 frame every 30 seconds at a resolution of $854 \times 480$.

\noindent \textbf{M2caiSeg} \cite{maqbool2020m2caiseg} dataset consists of 307 images extracted from 20 videos at the resolution of $774 \times 434$, annotated for the organs and different instruments present in the laparoscopic scene. Each video makes an annotation every 25 frames.

\noindent \textbf{AutoLaparoT3} \cite{wang2022autolaparo} comprises 1,800 laparoscopic frames at the resolution of $1,920 \times 1,080$. It is collected from 21 videos of the hysterectomy surgery, with annotations for the uterus and different surgical instrument segmentation tasks.

\noindent \textbf{Cholec80} \cite{twinanda2016endonet} contains 80 high-quality videos of gallbladder laparoscopic surgeries recorded at 25 frames per second, with Strasberg’s Critical View of Safety (CVS) criteria.

\noindent \textbf{CholecT50} \cite{nwoye2022rendezvous} is a dataset of endoscopic videos of laparoscopic cholecystectomy surgery, consisting of 45 videos from the Cholec80 \cite{twinanda2016endonet} dataset and 5 videos from the superset in-house Cholec120 \cite{nwoye2023}.

\noindent \textbf{EndoScapes-CVS201} \cite{murali2023endoscapes} includes 201 laparoscopic cholecystectomy videos. From these video clips, extracting 1 frame per second resulted in a dataset of 58,813 frames.

\noindent \textbf{Gynsurg} \cite{nasirihaghighi2025gynsurg} includes 15 pixel-level annotated laparoscopic hysterectomy videos for instrument and anatomy segmentation, totaling 12,362 frames at a resolution of $750 \times 480$. 

\begin{table*}[!t]
\centering
\small
\setlength\tabcolsep{3.2pt}
\caption{Comparison with state-of-the-arts on Surgical Instrument Segmentation.}
\adjustbox{width=1\textwidth,center}{\begin{tabular}{l|c|c|ccccccc|c|ccccccc}
\hline
\multirow{2}{*}{Methods} & \multirow{2}{*}{MP} & \multicolumn{8}{c|}{Dice (\%) $\uparrow$} & \multicolumn{8}{c}{HD (mm) $\downarrow$} \\
\cline{3-18}
& & P3 & C14 & C15 & C16 & C17 & C18 & C19 & C20 & P3 & C14 & C15 & C16 & C17 & C18 & C19 & C20 \\
\hline
nnUnet \cite{isensee2021nnu} & \multirow{6}{*}{\usym{2715}} & 67.27 & 58.43 & 35.90 & 65.40 & 60.01 & 67.13 & 52.68 & 61.75 & 234.67 & 678.92 & 512.38 & 198.54 & 287.63 & 245.82 & 356.74 & 223.45 \\
Surginet \cite{ni2022surginet} & & 68.21 & 61.74 & 75.00 & 69.53 & 58.58 & 15.28 & 65.44 & 58.92 & 218.93 & 623.48 & 189.76 & 176.29 & 312.57 & 598.34 & 298.67 & 287.53 \\
MSDE-Net \cite{yang2023msde} & & 69.84 & 63.29 & 72.56 & 68.54 & 64.93 & 48.67 & 75.68 & 67.38 & 203.58 & 587.26 & 215.93 & 184.72 & 256.48 & 423.65 & 213.89 & 245.67 \\
UniverSeg \cite{butoi2023universeg} & & 65.43 & 55.87 & 68.92 & 63.18 & 57.24 & 38.56 & 61.73 & 60.45 & 267.84 & 712.59 & 245.68 & 223.76 & 337.92 & 487.23 & 378.54 & 298.73 \\
SurgCSS \cite{zhao2025rethinking} & & 72.58 & 68.45 & 81.37 & 73.26 & 68.79 & 58.92 & 77.84 & 71.68 & 185.47 & 538.72 & 167.43 & 158.36 & 234.85 & 356.48 & 198.75 & 218.92 \\
Zig-RiR \cite{chen2025zig} & & 74.96 & 71.83 & 84.65 & 76.42 & 72.15 & 63.28 & 81.57 & 75.34 & 172.35 & 502.68 & 145.87 & 143.29 & 218.74 & 312.56 & 176.83 & 195.48 \\
\hline
SAM \cite{kirillov2023segment} & \multirow{6}{*}{\usym{1F5F8}} & 77.57 & 93.70 & 98.19 & 76.24 & 64.26 & 33.15 & 89.48 & 69.49 & 122.19 & 495.94 & 12.76 & 91.53 & 196.97 & 41.22 & 209.88 & 109.40 \\
SAM2 \cite{ravisam} & & 82.34 & 94.91 & 90.87 & 82.16 & 70.73 & 29.64 & 93.67 & 82.14 & 93.22 & 435.77 & 58.34 & 58.48 & 184.36 & 58.95 & 176.41 & 59.68 \\
MedSAM \cite{ma2024segment} & & 84.00 & 95.56 & 97.43 & 76.19 & 85.84 & 72.17 & 94.40 & 84.72 & 121.87 & 413.55 & 18.95 & \textbf{41.01} & 157.15 & 27.40 & 113.34 & 50.49 \\
SurgicalSAM \cite{yue2024surgicalsam} & & 83.75 & 95.38 & 96.82 & 81.54 & 84.26 & 65.83 & 92.85 & 86.47 & 98.63 & 421.82 & 23.47 & 48.72 & 168.54 & 32.68 & 128.76 & 55.93 \\
SurgSAM-2 \cite{liusurgical} & & 84.68 & 94.73 & 96.95 & 83.27 & 85.67 & 58.94 & 91.48 & 88.35 & 91.54 & 453.26 & 21.38 & 68.45 & 163.28 & 38.57 & 145.62 & 48.26 \\
ReSurgSAM2 \cite{liu2025resurgsam2} & & 85.21 & 94.14 & 97.38 & 84.35 & 86.48 & 47.46 & 89.30 & 90.38 & 88.40 & 476.31 & 19.67 & 103.65 & 156.95 & \textbf{22.63} & 131.93 & 52.70 \\
\hline
AutoSAM \cite{Shaharbany_2023_BMVC} & \multirow{3}{*}{\usym{2715}} & 75.42 & 91.58 & 96.73 & 74.86 & 62.47 & 30.54 & 87.62 & 67.83 & 134.68 & 523.47 & 24.89 & 103.57 & 218.93 & 67.84 & 232.45 & 125.68 \\
H-SAM \cite{cheng2024unleashing} & & 80.67 & 93.84 & 89.25 & 80.38 & 68.95 & 27.83 & 92.36 & 80.52 & 105.38 & 461.29 & 68.74 & 71.26 & 172.84 & 72.48 & 189.67 & 71.35 \\
ADA-SAM \cite{ward2025autoadaptive} & & 83.26 & 94.67 & 95.84 & 81.93 & 83.52 & 62.75 & 93.18 & 85.28 & 96.74 & 428.56 & 28.63 & 55.87 & 161.45 & 35.92 & 138.24 & 57.48 \\
\hline
LapFM & \usym{2715} & \textbf{90.81} & \textbf{97.60} & \textbf{99.17} & \textbf{89.75} & \textbf{88.08} & \textbf{97.59} & \textbf{95.54} & \textbf{98.02} & \textbf{84.99} & \textbf{106.32} & \textbf{5.19} & 85.98 & \textbf{155.42} & 145.05 & \textbf{97.62} & \textbf{4.56} \\
\hline
\end{tabular}}
\label{tab:instrument_comp}
\end{table*}

\subsubsection{Implementation Details}
We perform all experiments on two NVIDIA A100 GPUs using PyTorch. For fair comparisons, all segmentation methods are implemented with the same training settings and configurations. We utilize the pre-trained ViT-H \cite{kirillov2023segment} and Hirea-L \cite{ravisam} structures as the image encoder of SAM and SAM2 baselines, respectively. For our LapFM, we perform three iterations of the Confidence-driven Evolving Labeling with the selection ratios $t$ set as $70\%$. To mitigate potential data bias, we removed duplicate videos from the CholecT50 \cite{nwoye2022rendezvous}, Cholec80 \cite{twinanda2016endonet}, and EndoScapes-CVS201 \cite{murali2023endoscapes} datasets. We apply the optimizer using Adam with an initial learning rate of $1\times10^{-4}$ and use the exponential decay strategy to adjust the learning rate with a factor of 0.98. The loss coefficient $\lambda_1$ is set to 0.5. We split all datasets into the training, validation, and test sets as 7:1:2, and all images are resized to $1,024 \times 1,024$ during fine-tuning and evaluation stages, followed by the standard SAM preprocessing \cite{kirillov2023segment, ma2024segment}. The batch size and the training epoch are set to 16 and 200. In the comparison, we implement two modes of SAM-based methods, including the automatic mode without manual annotations and the prompt mode using the centroid of each surgical segmentation object as point prompts \cite{huang2024segment}, called Manual Prompt (MP). To perform a comprehensive evaluation, we select the Dice coefficient and Hausdorff distance (HD) as the main metrics in our experiments.

\subsection{Comparison on Anatomy Segmentation}

\begin{figure}[!t]
  \centering
  \includegraphics[width=1\linewidth]{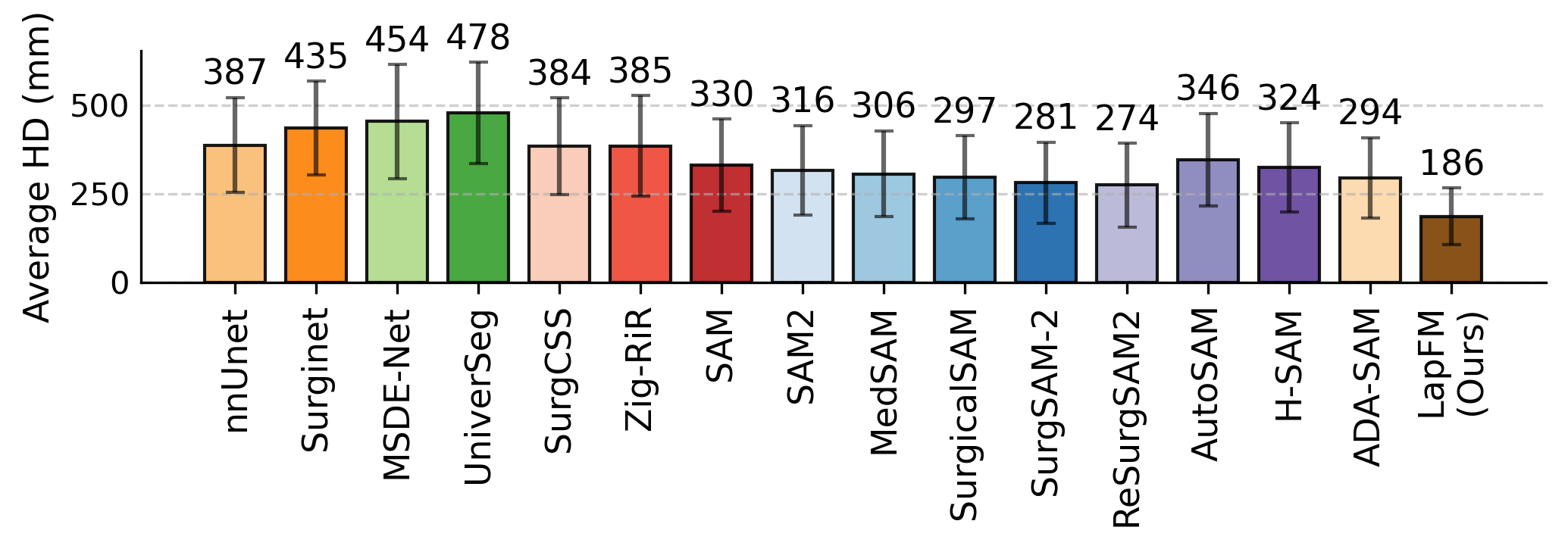}
  \caption{Comparison of average HD across different methods on anatomical structure segmentation. Our LapFM achieves substantially lower HD (186mm) compared to all baseline methods, demonstrating superior boundary localization precision.}
  \label{fig:hd_compar}
\end{figure}

To comprehensively evaluate the performance of LapFM on anatomical structure segmentation, we compare our method with state-of-the-art approaches across 12 anatomical categories on the combination of test sets \cite{hong2020cholecseg8k, carstens2023dresden, murali2023endoscapes, maqbool2020m2caiseg, wang2022autolaparo}. As shown in Table \ref{tab:anatomy_dice}, among methods without manual prompts, Zig-RiR \cite{chen2025zig} achieves 45.4\% average Dice, while among prompt-based methods, ReSurgSAM2 \cite{liu2025resurgsam2} demonstrates the strongest baseline performance with 68.5\% average Dice. In contrast, our LapFM substantially outperforms all baselines without requiring manual prompts, achieving 78.1\% average Dice with a remarkable 9.6\% absolute improvement over ReSurgSAM2. Notably, LapFM demonstrates exceptional performance on challenging fine-grained structures, particularly for Pancreas, where most methods struggle below 13.7\%, achieving 46.7\% Dice (3.4$\times$ improvement), and securing the best performance in 10 out of 12 anatomical categories. Furthermore, as illustrated in Fig.~\ref{fig:hd_compar}, LapFM achieves a substantially lower average Hausdorff distance of 186mm compared to all baseline methods, representing a 32.1\% reduction over the best competing method ReSurgSAM2 (274mm) and even more substantial improvements over auto-prompt methods like ADA-SAM \cite{ward2025autoadaptive} (385mm). These comparison results validate the superior generalization capability of our proposed LapFM with the Hierarchical Concept Evolving Pre-training paradigm for universal anatomy segmentation.

\begin{figure*}[!t]
  \centering
  \includegraphics[width=1\linewidth]{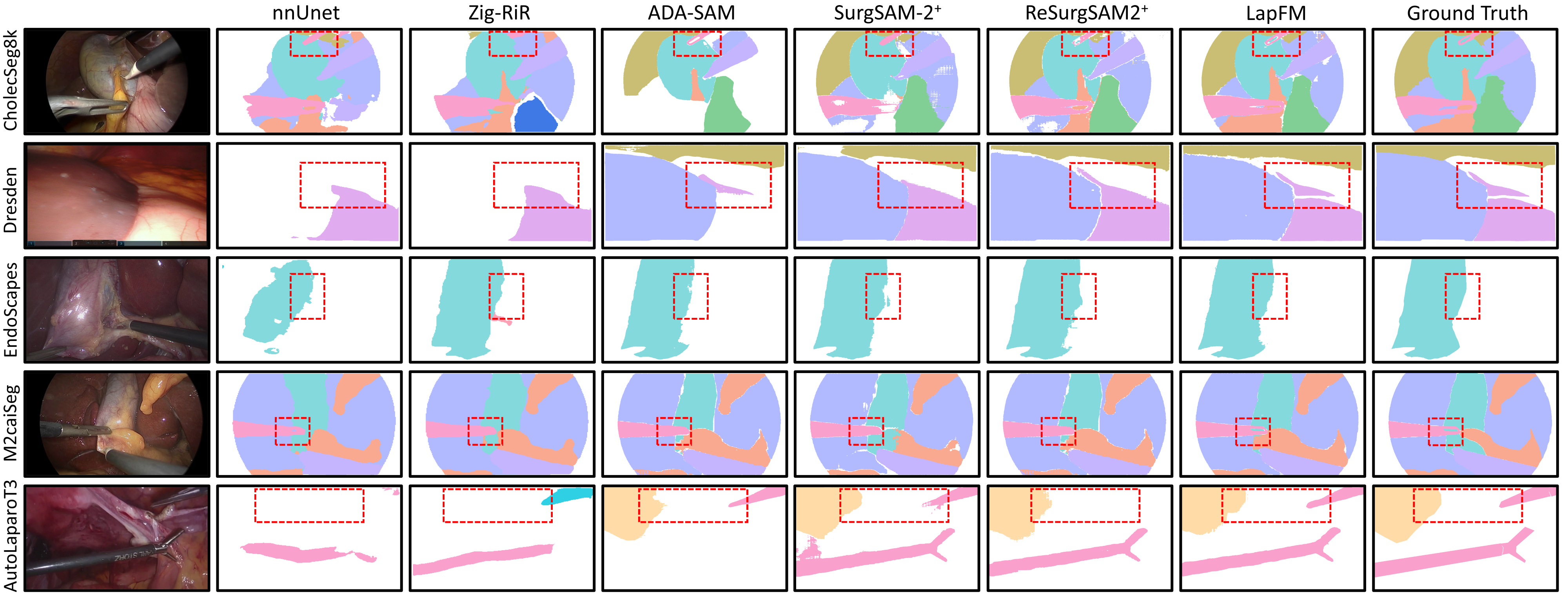}
  \caption{Visualization of laparoscopic surgical segmentation results across different methods. $+$ indicates surgical SAMs using center point prompts. Our LapFM exhibits the best results, segmenting instruments more accurately with precise boundary delineation while having fewer false positives and better robustness to occlusion and complex surgical scenarios.}
  \label{fig:model_compar}
\end{figure*}

\subsection{Comparison on Tissue Segmentation}

We further evaluate LapFM on tissue segmentation tasks on the combination of test sets \cite{hong2020cholecseg8k, carstens2023dresden, murali2023endoscapes, maqbool2020m2caiseg, wang2022autolaparo}, as shown in Table \ref{tab:tissue_comp}. Among methods without manual prompts, Zig-RiR \cite{chen2025zig} achieves the best baseline performance with 65.1\% average Dice. For prompt-based approaches, ReSurgSAM2 \cite{liu2025resurgsam2} demonstrates the strongest performance with 73.1\% average Dice, while the auto-prompting method ADA-SAM \cite{ward2025autoadaptive} reaches 65.7\%. Notably, Blood (C12) presents a particularly challenging segmentation task where most methods struggle: the best baseline ReSurgSAM2 achieves only 48.8\% Dice, with many methods failing to exceed 30\%. Our LapFM consistently surpasses all baselines across all tissue categories, achieving 94.8\% average Dice with a remarkable 21.7\% absolute improvement over ReSurgSAM2. LapFM excels particularly in the challenging Blood category with 98.4\% Dice and substantially lower HD (9.7mm vs. 80.4mm for ReSurgSAM2), representing a 7.3$\times$ reduction. These results validate the effectiveness of our hierarchical mask decoder in handling complex tissue boundaries and varying annotation granularities in laparoscopic surgical scenarios.

\subsection{Comparison on Instrument Segmentation}

To comprehensively evaluate the versatility of LapFM, we conduct extensive experiments on surgical instrument segmentation across 8 instrument categories on the combination of test sets \cite{hong2020cholecseg8k, carstens2023dresden, murali2023endoscapes, maqbool2020m2caiseg, wang2022autolaparo}. As shown in Table \ref{tab:instrument_comp}, among methods without manual prompts, Zig-RiR \cite{chen2025zig} achieves 72.5\% average Dice, while among prompt-based methods, MedSAM \cite{ma2024segment} reaches 86.3\% Dice, and among auto-prompting approaches, ADA-SAM \cite{ward2025autoadaptive} demonstrates the strongest baseline with 85.3\% average Dice. In contrast, LapFM consistently outperforms all baseline methods across all instrument types, achieving an impressive 94.6\% average Dice. Most remarkably, for the challenging Specimen Bag category (C18), where many methods struggle below 65\%, LapFM achieves 97.6\% Dice, a 35.0\% absolute improvement over the best baseline (62.8\%), while also achieving superior boundary precision with significantly lower Hausdorff distances across most categories. Moreover, we present qualitative comparisons in Fig.~\ref{fig:model_compar}, where our LapFM exhibits the best segmentation results, accurately delineating instrument boundaries with fewer false positives and superior handling of complex scenarios such as instrument occlusion. These results validate the effectiveness of our LapFM framework in handling complex surgical segmentation scenarios.

\begin{table}[!t]
\centering
\small
\setlength\tabcolsep{5pt}
\caption{Ablation Study of LapFM on Universal Laparoscopic Surgical Segmentation.}
\adjustbox{width=0.49\textwidth,center}{\begin{tabular}{cccc|ccc|ccc}
\hline
& \multirow{2}{*}{$T$} & \multirow{2}{*}{$F$} & \multirow{2}{*}{$E$} & \multicolumn{3}{c|}{Dice (\%) $\uparrow$} & \multicolumn{3}{c}{HD (mm) $\downarrow$} \\
\cline{5-10}
& & & & P1 & P2 & P3 & P1 & P2 & P3 \\
\hline
1 &  &  &  & 80.45 & 79.73 & 82.34  & 251.76 & 310.22 & 93.22\\
2 & \checkmark & & & 82.15 & 81.67 & 83.76 & 239.68 & 299.54 & 91.12\\
3 & & \checkmark & & 81.73 & 80.89 & 83.28 & 244.52 & 303.87 & 91.95\\
4 & & & \checkmark & 83.54 & 82.95 & 85.48 & 233.42 & 295.73 & 88.67\\
5 & \checkmark & \checkmark &  & 83.56 & 83.89 & 84.72 & 219.54 & 292.79 & 90.35\\
6 & \checkmark & & \checkmark & 85.36 & 88.54 & 88.97 & 197.38 & 284.62 & 86.84\\
7 & & \checkmark & \checkmark & 85.08 & 86.72 & 88.34 & 202.76 & 286.93 & 87.58\\
8 & \checkmark & \checkmark & \checkmark & \textbf{86.14} & \textbf{92.96} & \textbf{90.81} & \textbf{183.44} & \textbf{279.45} & \textbf{84.99} \\
\hline
\end{tabular}}
\label{tab:ablation}
\end{table}

\begin{figure}[!t]
  \centering
\includegraphics[width=0.47\textwidth]{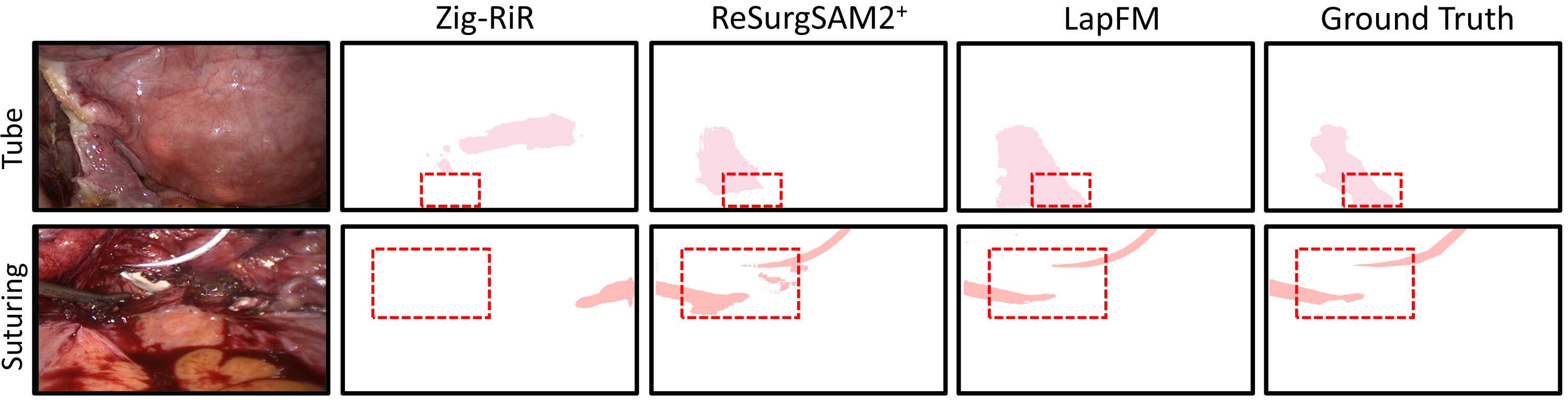}
\caption{Qualitative comparison of generalization capabilities on unseen surgical categories (Suturing and Tube) of the Gynsurg dataset.}
\label{fig:qualitative_zeroshot}
\end{figure}

\subsection{Ablation Study}

To investigate the effectiveness of the LCH $T$, LapFM architecture designs $F$, and Confidence-driven Evolving Labeling $E$, we conduct comprehensive ablation studies on universal laparoscopic surgical segmentation across the combination of test sets on five fully-annotated datasets, as illustrated in Table \ref{tab:ablation}. By removing the tailored modules from LapFM, we construct an original SAM2-based framework with a standard mask decoder as the ablation baseline. By separately introducing the LCH ($2^{nd}$ row), the LapFM architecture ($3^{rd}$ row), and the Confidence-driven Evolving Labeling ($4^{th}$ row), the performance demonstrates consistent improvements across all datasets. Notably, our LCH achieves a significant individual contribution, with substantial improvements averaging 1.67\% Dice across all protocols, highlighting its crucial role in dynamic category retrieval. We additionally investigate the effects of combined components. By comparing $6^{th}$ and $7^{th}$ rows with the $5^{th}$ row, configurations with Confidence-driven Evolving Labeling demonstrate significant gains, achieving an average Dice of 87.62\% and 86.71\%, respectively. On this basis, our complete LapFM framework ($8^{th}$ row) simultaneously adopts all three components to achieve the best performance. The full model attains optimal performance with an average Dice of 89.97\%, representing an average improvement of 9.13\% over the baseline. These comprehensive results validate that the tailored LCH, LapFM architecture designs, and Confidence-driven Evolving Labeling collectively contribute to the superior performance of our framework.

\begin{figure}[!t]
  \centering
\includegraphics[width=0.49\textwidth]{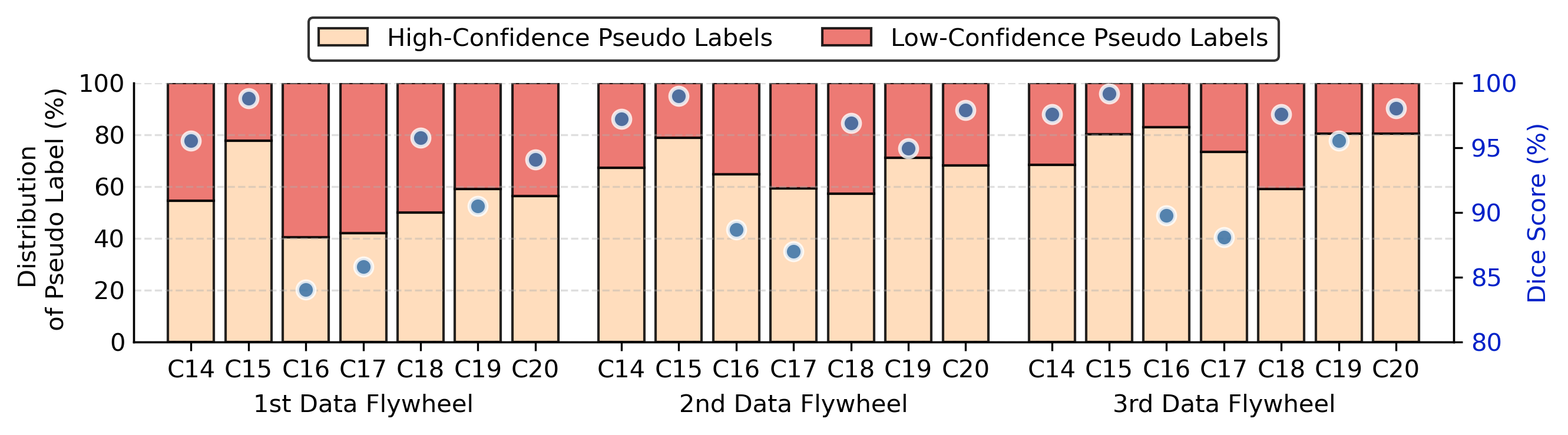}
\caption{Comparison of pseudo-label confidence evolution and model performance across iterations in surgical instrument segmentation. The Confidence-driven Evolving Labeling progressively increases the proportion of high-confidence pseudo-labels while reducing low-confidence ones, with enhanced model performance.}
\label{fig:data_flywheel}
\end{figure}

\begin{figure*}[!t]
  \centering
\includegraphics[width=1\textwidth]{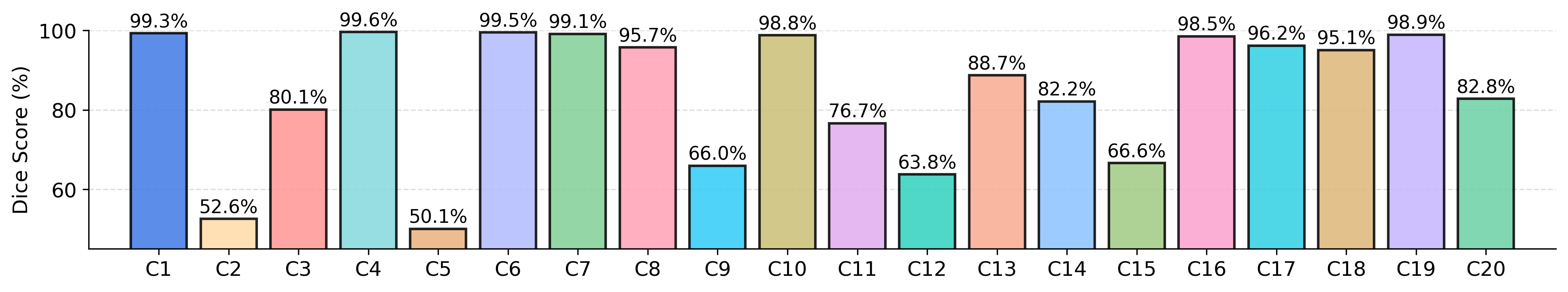}
\caption{Comparison of surgical expert annotations with pseudo-labels across 20 laparoscopic surgical categories. This expert validation confirms that our Confidence-driven Evolving Labeling successfully generates reliable pseudo-labels for the majority of surgical categories.}
\label{fig:pseudo_comparison}
\end{figure*}

\begin{table}[!t]
\centering
\small
\setlength\tabcolsep{3pt}
\caption{Generalization Comparison with State-of-the-art Methods on the unseen GynSurg dataset.}
\adjustbox{width=0.48\textwidth,center}{\begin{tabular}{l|c|cccccc}
\hline
\multirow{2}{*}{Methods} & \multirow{2}{*}{MP} & \multicolumn{6}{c}{Dice (\%)} \\
\cline{3-8}
& & C2 & C14 & C16 & C17 & C19 & C20 \\
\hline
nnUnet \cite{isensee2021nnu} & \multirow{6}{*}{\usym{2715}} & 16.61 & 3.51 & 12.77 & 0.90 & 3.18 & 1.45 \\
Surginet \cite{ni2022surginet} & & 14.78 & 5.23 & 9.91 & 8.11 & 2.73 & 6.84 \\
MSDE-Net \cite{yang2023msde} & & 20.03 & 8.67 & 18.73 & 16.44 & 16.89 & 12.38 \\
UniverSeg \cite{butoi2023universeg} & & 23.67 & 6.94 & 22.97 & 3.17 & 28.04 & 8.52 \\
SurgCSS \cite{zhao2025rethinking} & & 18.80 & 11.45 & 19.72 & 8.86 & 13.42 & 15.27 \\
Zig-RiR \cite{chen2025zig} & & 31.62 & 14.83 & 17.71 & 12.09 & 25.79 & 18.94 \\
\hline
SAM \cite{kirillov2023segment} & \multirow{6}{*}{\usym{1F5F8}} & 28.45 & 24.73 & 26.82 & 19.54 & 32.67 & 21.38 \\
SAM2 \cite{ravisam} & & 32.68 & 28.95 & 31.47 & 24.83 & 38.92 & 27.65 \\
MedSAM \cite{ma2024segment} & & 35.92 & 32.47 & 34.68 & 28.95 & 42.83 & 31.74 \\
SurgicalSAM \cite{yue2024surgicalsam} & & 38.74 & 35.68 & 37.93 & 32.47 & 46.25 & 35.82 \\
SurgSAM-2 \cite{liusurgical} & & 42.35 & 39.84 & 41.68 & 36.92 & 51.47 & 40.28 \\
ReSurgSAM2 \cite{liu2025resurgsam2} & & 46.83 & \textbf{43.52} & 45.94 & 41.38 & 55.84 & 44.67 \\
\hline
AutoSAM \cite{Shaharbany_2023_BMVC} & \multirow{3}{*}{\usym{2715}} & 26.73 & 22.48 & 24.95 & 17.82 & 30.54 & 19.67 \\
H-SAM \cite{cheng2024unleashing} & & 30.84 & 26.92 & 29.73 & 22.68 & 36.47 & 25.83 \\
ADA-SAM \cite{ward2025autoadaptive} & & 37.56 & 34.28 & 36.85 & 31.74 & 45.38 & 34.92 \\
\hline
LapFM & \usym{2715} & \textbf{61.23} & 35.51 & \textbf{67.32} & \textbf{47.28} & \textbf{58.16} & \textbf{48.17} \\
\hline
\end{tabular}}
\label{tab_zeroshot}
\end{table}

\subsection{Comparison on Generalization Capability}

To further validate the generalization capability of our LapFM, we conduct a comparison on the unseen GynSurg dataset containing 6 surgical categories. As shown in Table \ref{tab_zeroshot}, compared to the best-performing baseline Zig-RiR \cite{chen2025zig}, LapFM achieves dramatic improvements with an average Dice of 34.45\%. For individual categories, LapFM demonstrates exceptional performance on Uterus (61.23\% vs. 31.62\%) and Grasper (67.32\% vs. 17.71\%), while maintaining robust segmentation quality across all surgical entities. While SAM-based foundation models with manual prompts generally achieve better performance than conventional methods, LapFM significantly surpasses even the most advanced surgical SAM variants. Compared to ReSurgSAM2 \cite{liu2025resurgsam2}, the best-performing manual-prompt baseline achieving 46.36\% average Dice, LapFM demonstrates an 8.25\% improvement without requiring any manual intervention. Our method reveals statistically significant performance advantages (P-value $<$ 0.001) across all surgical SAM variants. 

We further conduct additional qualitative evaluation on two unseen surgical categories: Suturing and Tube, as visualized in Fig. \ref{fig:qualitative_zeroshot}. Classical methods like Zig-RiR \cite{chen2025zig}, constrained by predefined category spaces, fail to adapt to these new categories. While SAM-based models can segment unseen categories, they critically depend on labor-intensive manual prompts for each target. In contrast, LapFM leverages confidence-guided adaptive traversal through the LCH to dynamically match unseen entities to appropriate granularity levels without manual intervention. These comprehensive comparisons demonstrate the superior generalization capability of our LapFM framework. The significant improvements across diverse unseen surgical categories, anatomical structures, and granularity levels confirm the significant potential of LapFM for deployment in the real-world clinical scenarios and emerging surgical procedures, where the expert-based manual annotation or prompting is impractical.

\subsection{Professional Assessment of Confidence-driven Evolving Labeling}

To further analyze the effectiveness of our Confidence-driven Evolving Labeling, we visualize the evolution of pseudo-label quality and model performance across iterations in Fig.~\ref{fig:data_flywheel}. The results demonstrate a clear progressive refinement pattern: as the iterations advance, the proportion of high-confidence pseudo-labels steadily increases, while the proportion of low-confidence samples simultaneously decreases. This progressive confidence distribution shift directly correlates with substantial performance improvements, where the segmentation accuracy exhibits continuous enhancement. To further validate the efficiency of pseudo-labels generated by our Confidence-driven Evolving Labeling, we randomly sample 3 instances per category across all 20 surgical categories. These samples are manually annotated by an experienced laparoscopic surgeon. We then quantitatively assess the annotation quality, as illustrated in Fig. \ref{fig:pseudo_comparison}. The overall average Dice score across all 20 categories reaches 80.07\%, demonstrating acceptable annotation quality. These statistical results validate the effectiveness of the proposed Confidence-driven Evolving Labeling, ensuring annotation quality while maximizing data utilization across diverse surgical scenarios with varying annotation granularities.

\section{Conclusion}
In this work, we present LapFM, a laparoscopic foundation model built via Hierarchical Concept Evolving Pre-training. We construct a LCH that unifies diverse surgical entities into a scalable taxonomy, enabling granularity-adaptive segmentation through a hierarchical mask decoder with explicit parent-child dependencies. Additionally, we propose a Confidence-driven Evolving Labeling that progressively transforms massive unlabeled surgical images into high-quality supervision, yielding LapBench-114K comprising 114K image-mask pairs across 20 surgical categories. Extensive experiments demonstrate that our LapFM framework consistently outperforms state-of-the-art methods, achieving superior generalization capabilities on unseen laparoscopic surgical scenes.

\balance
\bibliographystyle{IEEEtran}
\bibliography{ref}

\begin{thebibliography}{10}
\providecommand{\url}[1]{#1}
\csname url@samestyle\endcsname
\providecommand{\newblock}{\relax}
\providecommand{\bibinfo}[2]{#2}
\providecommand{\BIBentrySTDinterwordspacing}{\spaceskip=0pt\relax}
\providecommand{\BIBentryALTinterwordstretchfactor}{4}
\providecommand{\BIBentryALTinterwordspacing}{\spaceskip=\fontdimen2\font plus
\BIBentryALTinterwordstretchfactor\fontdimen3\font minus \fontdimen4\font\relax}
\providecommand{\BIBforeignlanguage}[2]{{%
\expandafter\ifx\csname l@#1\endcsname\relax
\typeout{** WARNING: IEEEtran.bst: No hyphenation pattern has been}%
\typeout{** loaded for the language `#1'. Using the pattern for}%
\typeout{** the default language instead.}%
\else
\language=\csname l@#1\endcsname
\fi
#2}}
\providecommand{\BIBdecl}{\relax}
\BIBdecl

\bibitem{twinanda2016endonet}
A.~P. Twinanda, S.~Shehata, D.~Mutter, J.~Marescaux, M.~De~Mathelin, and N.~Padoy, ``Endonet: a deep architecture for recognition tasks on laparoscopic videos,'' \emph{IEEE Trans. Med. Imaging}, vol.~36, no.~1, pp. 86--97, 2016.

\bibitem{kolbinger2023anatomy}
F.~R. Kolbinger, F.~M. Rinner, A.~C. Jenke, M.~Carstens, S.~Krell, S.~Leger, M.~Distler, J.~Weitz, S.~Speidel, and S.~Bodenstedt, ``Anatomy segmentation in laparoscopic surgery: comparison of machine learning and human expertise--an experimental study,'' \emph{Int. J. Surg.}, vol. 109, no.~10, pp. 2962--2974, 2023.

\bibitem{ronneberger2015u}
O.~Ronneberger, P.~Fischer, and T.~Brox, ``U-net: Convolutional networks for biomedical image segmentation,'' in \emph{MICCAI}.\hskip 1em plus 0.5em minus 0.4em\relax Springer, 2015, pp. 234--241.

\bibitem{chen2024transunet}
J.~Chen, J.~Mei, X.~Li, Y.~Lu, Q.~Yu, Q.~Wei, X.~Luo, Y.~Xie, E.~Adeli, Y.~Wang \emph{et~al.}, ``Transunet: Rethinking the u-net architecture design for medical image segmentation through the lens of transformers,'' \emph{Med. Image Anal.}, vol.~97, p. 103280, 2024.

\bibitem{maack2024efficient}
L.~Maack, F.~Behrendt, D.~Bhattacharya, S.~Latus, and A.~Schlaefer, ``Efficient anatomy segmentation in laparoscopic surgery using multi-teacher knowledge distillation,'' in \emph{MIDL}, 2024.

\bibitem{tomar2025effective}
P.~Tomar, A.~Parikh, P.~Feodorovici, J.~Arensmeyer, H.~Matthaei, C.~Bauckhage, H.~Schneider, and R.~Sifa, ``Effective disjoint representational learning for anatomical segmentation,'' in \emph{MIDL}, 2025.

\bibitem{chen2023surgnet}
J.~Chen, M.~Li, H.~Han, Z.~Zhao, and X.~Chen, ``Surgnet: Self-supervised pretraining with semantic consistency for vessel and instrument segmentation in surgical images,'' \emph{IEEE Trans. Med. Imaging}, vol.~43, no.~4, pp. 1513--1525, 2023.

\bibitem{yang2023msde}
Y.~Lei, Y.~Gu, G.~Bian, and Y.~Liu, ``Msde-net: A multi-scale dual-encoding network for surgical instrument segmentation,'' \emph{IEEE J. Biomed. Health Inform.}, vol.~28, no.~7, pp. 4072--4083, 2023.

\bibitem{zhao2025rethinking}
S.~Zhao, L.~Bai, K.~Yuan, F.~Li, J.~Yu, W.~Dong, G.~Wang, M.~Islam, N.~Padoy, N.~Navab \emph{et~al.}, ``Rethinking data imbalance in class incremental surgical instrument segmentation,'' \emph{Med. Image Anal.}, p. 103728, 2025.

\bibitem{carstens2023dresden}
M.~Carstens, F.~M. Rinner, S.~Bodenstedt, A.~C. Jenke, J.~Weitz, M.~Distler, S.~Speidel, and F.~R. Kolbinger, ``The dresden surgical anatomy dataset for abdominal organ segmentation in surgical data science,'' \emph{Sci. Data}, vol.~10, no.~1, pp. 1--8, 2023.

\bibitem{hong2020cholecseg8k}
W.-Y. Hong, C.-L. Kao, Y.-H. Kuo, J.-R. Wang, W.-L. Chang, and C.-S. Shih, ``Cholecseg8k: a semantic segmentation dataset for laparoscopic cholecystectomy based on cholec80,'' \emph{arXiv preprint arXiv:2012.12453}, 2020.

\bibitem{wang2022autolaparo}
Z.~Wang, B.~Lu, Y.~Long, F.~Zhong, T.-H. Cheung, Q.~Dou, and Y.~Liu, ``Autolaparo: A new dataset of integrated multi-tasks for image-guided surgical automation in laparoscopic hysterectomy,'' in \emph{MICCAI}.\hskip 1em plus 0.5em minus 0.4em\relax Springer, 2022, pp. 486--496.

\bibitem{kirillov2023segment}
A.~Kirillov, E.~Mintun, N.~Ravi, H.~Mao, C.~Rolland, L.~Gustafson, T.~Xiao, S.~Whitehead, A.~C. Berg, W.-Y. Lo \emph{et~al.}, ``Segment anything,'' in \emph{ICCV}, 2023, pp. 4015--4026.

\bibitem{ravisam}
N.~Ravi, V.~Gabeur, Y.-T. Hu, R.~Hu, C.~Ryali, T.~Ma, H.~Khedr, R.~R{\"a}dle, C.~Rolland, L.~Gustafson \emph{et~al.}, ``Sam 2: Segment anything in images and videos,'' in \emph{ICLR}, 2024.

\bibitem{yue2024surgicalsam}
W.~Yue, J.~Zhang, K.~Hu, Y.~Xia, J.~Luo, and Z.~Wang, ``Surgicalsam: Efficient class promptable surgical instrument segmentation,'' in \emph{AAAI}, vol.~38, no.~7, 2024, pp. 6890--6898.

\bibitem{liusurgical}
H.~Liu, E.~Zhang, J.~Wu, M.~Hong, and Y.~Jin, ``Surgical sam 2: Real-time segment anything in surgical video by efficient frame pruning,'' in \emph{NeurIPS Workshop AIM-FM}, 2024.

\bibitem{kamtam2025surgisam2}
D.~N. Kamtam, J.~B. Shrager, S.~D. Malla, X.~Wang, N.~Lin, J.~J. Cardona, S.~Yeung-Levy, and C.~Hu, ``Surgisam2: fine-tuning a foundational model for surgical video anatomy segmentation and detection,'' \emph{arXiv preprint arXiv:2503.03942}, 2025.

\bibitem{liu2025resurgsam2}
H.~Liu, M.~Gao, X.~Luo, Z.~Wang, G.~Qin, J.~Wu, and Y.~Jin, ``Resurgsam2: Referring segment anything in surgical video via credible long-term tracking,'' \emph{MICCAI}, 2025.

\bibitem{sheng2024surgical}
Y.~Sheng, S.~Bano, M.~J. Clarkson, and M.~Islam, ``Surgical-desam: decoupling sam for instrument segmentation in robotic surgery,'' \emph{Int. J. Comput. Assist. Radiol. Surg.}, vol.~19, no.~7, pp. 1267--1271, 2024.

\bibitem{huang2024segment}
Y.~Huang, X.~Yang, L.~Liu, H.~Zhou, A.~Chang, X.~Zhou, R.~Chen, J.~Yu, J.~Chen, C.~Chen \emph{et~al.}, ``Segment anything model for medical images?'' \emph{Med. Image Anal.}, vol.~92, p. 103061, 2024.

\bibitem{wu2025medical}
J.~Wu, Z.~Wang, M.~Hong, W.~Ji, H.~Fu, Y.~Xu, M.~Xu, and Y.~Jin, ``Medical sam adapter: Adapting segment anything model for medical image segmentation,'' \emph{Med. Image Anal.}, vol. 102, p. 103547, 2025.

\bibitem{yan2025samed}
Z.~Yan, S.~Song, D.~Song, Y.~Li, R.~Zhou, W.~Sun, Z.~Chen, S.~Kim, H.~Ren, T.~Liu \emph{et~al.}, ``Samed-2: Selective memory enhanced medical segment anything model,'' \emph{MICCAI}, 2025.

\bibitem{xu2025lightsam}
Q.~Xu, J.~Li, X.~He, C.~Li, F.~B. Tesema, W.~Duan, Z.~Chen, R.~Qu, J.~M. Garibaldi, and C.~W. Chen, ``De-lightsam: Modality-decoupled lightweight sam for generalizable medical segmentation,'' \emph{IEEE Trans. Circuits Syst. Video Technol.}, 2025.

\bibitem{Shaharbany_2023_BMVC}
S.~Tal, D.~Aviad, G.~Raja, and W.~Lior, ``Autosam: Adapting sam to medical images by overloading the prompt encoder,'' in \emph{BMVC}.\hskip 1em plus 0.5em minus 0.4em\relax BMVA, 2023.

\bibitem{cheng2024unleashing}
Z.~Cheng, Q.~Wei, H.~Zhu, Y.~Wang, L.~Qu, W.~Shao, and Y.~Zhou, ``Unleashing the potential of sam for medical adaptation via hierarchical decoding,'' in \emph{CVPR}, 2024, pp. 3511--3522.

\bibitem{ward2025autoadaptive}
T.~Ward, M.~K. Owen, O.~Coleman, B.~Noehren, and A.-A.-Z. Imran, ``Autoadaptive medical segment anything model,'' \emph{arXiv preprint arXiv:2507.01828}, 2025.

\bibitem{nwoye2022rendezvous}
C.~I. Nwoye, T.~Yu, C.~Gonzalez, B.~Seeliger, P.~Mascagni, D.~Mutter, J.~Marescaux, and N.~Padoy, ``Rendezvous: Attention mechanisms for the recognition of surgical action triplets in endoscopic videos,'' \emph{Med. Image Anal.}, vol.~78, p. 102433, 2022.

\bibitem{zhou2021models}
Z.~Zhou, V.~Sodha, J.~Pang, M.~B. Gotway, and J.~Liang, ``Models genesis,'' \emph{Med. Image Anal.}, vol.~67, p. 101840, 2021.

\bibitem{ye2023uniseg}
Y.~Ye, Y.~Xie, J.~Zhang, Z.~Chen, and Y.~Xia, ``Uniseg: A prompt-driven universal segmentation model as well as a strong representation learner,'' in \emph{MICCAI}.\hskip 1em plus 0.5em minus 0.4em\relax Springer, 2023, pp. 508--518.

\bibitem{murali2023endoscapes}
A.~Murali, D.~Alapatt, P.~Mascagni, A.~Vardazaryan, A.~Garcia, N.~Okamoto, G.~Costamagna, D.~Mutter, J.~Marescaux, B.~Dallemagne \emph{et~al.}, ``The endoscapes dataset for surgical scene segmentation, object detection, and critical view of safety assessment: Official splits and benchmark,'' \emph{arXiv preprint arXiv:2312.12429}, 2023.

\bibitem{wang2023sam}
A.~Wang, M.~Islam, M.~Xu, Y.~Zhang, and H.~Ren, ``Sam meets robotic surgery: an empirical study on generalization, robustness and adaptation,'' in \emph{MICCAI}.\hskip 1em plus 0.5em minus 0.4em\relax Springer, 2023, pp. 234--244.

\bibitem{nasirihaghighi2025gynsurg}
S.~Nasirihaghighi, N.~Ghamsarian, L.~Peschek, M.~Munari, H.~Husslein, R.~Sznitman, and K.~Schoeffmann, ``Gynsurg: A comprehensive gynecology laparoscopic surgery dataset,'' in \emph{ACM MM}, 2025, pp. 13\,141--13\,147.

\bibitem{liu2024swin}
J.~Liu, H.~Yang, H.-Y. Zhou, L.~Yu, Y.~Liang, Y.~Yu, S.~Zhang, H.~Zheng, and S.~Wang, ``Swin-umamba†: Adapting mamba-based vision foundation models for medical image segmentation,'' \emph{IEEE Trans. Med. Imaging}, 2024.

\bibitem{chen2025zig}
T.~Chen, X.~Zhou, Z.~Tan, Y.~Wu, Z.~Wang, Z.~Ye, T.~Gong, Q.~Chu, N.~Yu, and L.~Lu, ``Zig-rir: Zigzag rwkv-in-rwkv for efficient medical image segmentation,'' \emph{IEEE Trans. Med. Imaging}, 2025.

\bibitem{ni2022surginet}
Z.-L. Ni, X.-H. Zhou, G.-A. Wang, W.-Q. Yue, Z.~Li, G.-B. Bian, and Z.-G. Hou, ``Surginet: Pyramid attention aggregation and class-wise self-distillation for surgical instrument segmentation,'' \emph{Med. Image Anal.}, vol.~76, p. 102310, 2022.

\bibitem{yang2022tmf}
L.~Yang, Y.~Gu, G.~Bian, and Y.~Liu, ``Tmf-net: A transformer-based multiscale fusion network for surgical instrument segmentation from endoscopic images,'' \emph{IEEE Transactions on Instrumentation and Measurement}, vol.~72, pp. 1--15, 2022.

\bibitem{ma2024segment}
J.~Ma, Y.~He, F.~Li, L.~Han, C.~You, and B.~Wang, ``Segment anything in medical images,'' \emph{Nat. Commun.}, vol.~15, no.~1, p. 654, 2024.

\bibitem{chen2024asi}
Z.~Chen, Z.~Zhang, W.~Guo, X.~Luo, L.~Bai, J.~Wu, H.~Ren, and H.~Liu, ``Asi-seg: Audio-driven surgical instrument segmentation with surgeon intention understanding,'' in \emph{IROS}.\hskip 1em plus 0.5em minus 0.4em\relax IEEE, 2024, pp. 13\,773--13\,779.

\bibitem{luddecke2022image}
T.~L{\"u}ddecke and A.~Ecker, ``Image segmentation using text and image prompts,'' in \emph{CVPR}, 2022, pp. 7086--7096.

\bibitem{liu2023clip}
J.~Liu, Y.~Zhang, J.-N. Chen, J.~Xiao, Y.~Lu, B.~A~Landman, Y.~Yuan, A.~Yuille, Y.~Tang, and Z.~Zhou, ``Clip-driven universal model for organ segmentation and tumor detection,'' in \emph{ICCV}, 2023, pp. 21\,152--21\,164.

\bibitem{isensee2021nnu}
F.~Isensee, P.~F. Jaeger, S.~A. Kohl, J.~Petersen, and K.~H. Maier-Hein, ``nnu-net: a self-configuring method for deep learning-based biomedical image segmentation,'' \emph{Nature Methods}, vol.~18, no.~2, pp. 203--211, 2021.

\bibitem{butoi2023universeg}
V.~I. Butoi, J.~J.~G. Ortiz, T.~Ma, M.~R. Sabuncu, J.~Guttag, and A.~V. Dalca, ``Universeg: Universal medical image segmentation,'' in \emph{ICCV}, 2023, pp. 21\,438--21\,451.

\bibitem{mascagni2025endoscapes}
P.~Mascagni, D.~Alapatt, A.~Murali, A.~Vardazaryan, A.~Garcia, N.~Okamoto, G.~Costamagna, D.~Mutter, J.~Marescaux, B.~Dallemagne \emph{et~al.}, ``Endoscapes, a critical view of safety and surgical scene segmentation dataset for laparoscopic cholecystectomy,'' \emph{Sci. Data}, vol.~12, no.~1, p. 331, 2025.

\bibitem{maqbool2020m2caiseg}
S.~Maqbool, A.~Riaz, H.~Sajid, and O.~Hasan, ``m2caiseg: Semantic segmentation of laparoscopic images using convolutional neural networks,'' \emph{arXiv preprint arXiv:2008.10134}, 2020.

\bibitem{nwoye2023}
C.~I. Nwoye and N.~Padoy, ``Data splits and metrics for method benchmarking on surgical action triplet datasets,'' in \emph{arXiv}, 2023.

\end{thebibliography}

\end{document}